\documentclass[journal]{IEEEtran}
\usepackage{amsmath,amsfonts}
\usepackage{amssymb}
\usepackage{algorithmic}
\usepackage{algorithm}
\usepackage{array}
\usepackage[caption=false,font=scriptsize,labelfont=sf,textfont=sf]{subfig}
\usepackage{textcomp}
\usepackage{stfloats}
\usepackage{multirow}
\usepackage{url}
\usepackage{verbatim}
\usepackage{graphicx}
\usepackage{adjustbox}
\usepackage{tikz}
\usetikzlibrary{arrows.meta,positioning,fit,calc,shapes.geometric}
\usepackage{cite}
\author{
Alireza Dastmalchi Saei,~\IEEEmembership{Member,~IEEE,} and Shervin Rahimzadeh Arashloo,~\IEEEmembership{Member,~IEEE,}%
\thanks{A. Dastmalchi Saei and S. R. Arashloo are with the Department of Computer Engineering, Bilkent University, 06800 Ankara, T\"urkiye. Corresponding author: S.R. Arashloo (e-mail: sh.rahimzadeh@hotmail.com).}
}

\begin{document}

\title{Deep Convolutional Large-Margin $\ell_p$-SVDD for Visual Anomaly Detection}

\maketitle

\begin{tikzpicture}[remember picture,overlay]
\node[
    anchor=north,
    align=center,
    text width=0.94\paperwidth,
    font=\fontsize{7}{8}\selectfont
] at ([yshift=-8pt]current page.north)
{
This work has been submitted to the IEEE for possible publication.
Copyright may be transferred without notice, after which this version
may no longer be accessible.
};
\end{tikzpicture}

\begin{abstract}
Visual anomaly detection requires adaptive representations and reliable decision boundaries, particularly when anomalous training samples are scarce and class distributions are highly imbalanced. Classical kernel-based methods yield principled geometric decision regions but typically operate on fixed features, while deep detectors learn task-specific representations but often fail to provide an explicit margin-aware kernel boundary. In this study, we propose DLM-SVDD, a deep large-margin novelty-detection framework that jointly learns convolutional features and an explicit kernel-based decision boundary. By drawing on the large-margin $\ell_p$-Support Vector Data Description ($\ell_p$-SVDD) approach, the proposed method performs explicit margin maximization and nonlinear slack penalization while adapting the representation to the target task. To train the proposed model, we present an optimization scheme that alternates between a Frank--Wolfe--based update of the convex dual boundary and a CNN update step operating on a smooth margin-violation loss induced by the recovered boundary. To improve scalability, we analyze the efficiency--accuracy trade-offs for different kernel approximation strategies, deriving practical propositions for large-scale anomaly detection.

Extensive experiments on multiple standard benchmarks show consistent performance improvements over the baseline and strong overall performance compared with state-of-the-art methods while illustrating that the proposed joint representation--boundary learning scheme remains effective under severe imbalanced class distributions.
\end{abstract}


\section{Introduction}
\IEEEPARstart{V}{isual} anomaly detection has become a core component of reliable image processing pipelines, where the objective is to detect observations that depart from a nominal data distribution despite limited access to representative abnormal examples. Such departures may arise from images belonging to unexpected visual categories, rare semantic patterns, or unexpected modes of data distribution. This inherent nature of anomaly detection makes the problem challenging as the model must learn a discriminative description of normality while remaining sensitive to deviations that may be scarce, diverse,  only partially observed during training, or completely absent. The difficulty is further amplified when the available training data are highly imbalanced, since dominant classes or modes can bias the learned representation and distort the resulting decision boundary. These observations motivate anomaly-detection methods that combine adaptive visual representations with reliable decision geometry, while remaining stable and scalable under limited supervision and class-imbalanced training scenarios.

An important design consideration in anomaly detection is how to couple \emph{representation learning} with explicit \emph{decision geometry}. Recent approaches benefiting from large-scale pretraining and feature adaptation often provide strong transferability~\cite{reiss2021panda,reiss2023msc}, but they can be sensitive to the choice of scoring rule and may not impose a principled large-margin separation in the learned decision space. Conversely, and as a completely different avenue, classical kernel methods offer well-defined decision boundaries, but may not perform on par with deep networks when used with fixed, non-adaptive features for complex imagery and may become computationally demanding when exact kernel matrices are utilized for large training sets. This dilemma suggests that for high accuracy, anomaly detection methods may concurrently benefit from: (i) features that adapt to the manifold of the target domain; and (ii) a decision function whose geometry is optimized via a stable and theoretically grounded criterion. Despite some earlier attempts \cite{MOHAMMAD2024127246} achieving these two objectives simultaneously in a unified framework remains a challenge and largely unexplored, particularly when training data is scarce, class distributions are imbalanced, or the feature space is high-dimensional.

As a de facto standard for one-class classification and novelty detection, the Support Vector Data Description (SVDD) approach \cite{tax2004support} provides a canonical formulation of one-class learning by enclosing normal data within a minimal-volume hypersphere while allowing violations through slack variables. Deep extensions~\cite{ruff2018deep,Ruff2020Deep, hojjati2023dasvdd,xing2024dmsvdd} have improved the effectiveness of the approach, yet many of these methods simplify the original objective or rely on surrogate losses that weaken the connection to margin-based learning. On the other hand, the recently introduced large-margin $\ell_p$-SVDD~\cite{arashloo2024large} framework explicitly maximizes a separation margin between target and non-target objects, applies nonlinear ($\ell_p$-norm) penalization to constraint violations, and admits efficient optimization through a Frank--Wolfe procedure on the convex dual objective function. Yet, the method suffers from two limitations. First, it operates on \emph{fixed} features, thereby decoupling boundary learning from feature learning, a procedure which may be considered suboptimal for high-dimensional visual data analysis. Second, exact kernel construction and repeated kernel operations may become computationally too expensive as the training set grows, motivating scalable approximations that preserve the structure of the kernel boundary while keeping computational cost within affordable budgets.

Motivated by the aforementioned observations, this study presents an end-to-end learning approach by integrating a trainable convolutional backbone with the large-margin $\ell_p$-SVDD objective while preserving its margin-driven geometry and optimization structure. The resulting method, which we refer to as DLM-SVDD (Deep Large Margin Support Vector Data Description), conducts two complementary updates within a stable alternating scheme: (1) a boundary-fitting step that optimizes the dual large-margin $\ell_p$-SVDD objective function via a Frank--Wolfe-based approach given the current deep features; and (2), a network update step that optimizes a smooth margin-violation loss to minimize constraint violation errors. Importantly, as the gradient signal in the network update step would be largely shaped by margin-violating samples, representation learning will be more focused on the corresponding decision regions of the feature space, thus avoiding optimization distraction induced by other parts. Such a joint design scheme preserves the theoretical rigor and margin maximization properties of the large-margin $\ell_p$-SVDD methodology while enabling adaptive feature learning in a unified framework. To address scalability for large-scale training, we further equip the proposed approach with a family of kernel approximation backends that reduce the cost of relevant kernel matrix operations while preserving the general structure of the problem, enabling the framework to scale to larger datasets while preserving competitive detection performance.

In summary, we make the following contributions:
\begin{itemize}
    \item We introduce the DLM-SVDD approach, which jointly learns convolutional representations and an explicit large-margin $\ell_p$-SVDD boundary, extending the fixed-feature framework of~\cite{arashloo2024large} to adaptive deep CNN features.

    \item We propose a stable alternating optimization procedure combining a bespoke Frank--Wolfe-based dual update mechanism for decision boundary estimation with a margin-aware primal loss update step for CNN parameter learning while providing convergence and stability analyses of the proposed alternating scheme.
    
    \item Within the proposed approach, we analyze several kernel matrix approximation schemes, including Nystr\"om, RFF, QMC-RFF, ORF, SORF, Fastfood, and RPCholesky~\cite{williams2000nystrom,rahimi2007random, avron2016qmc,yu2016orf,le2013fastfood,chen2025randomly} to enable scaling to large training corpora, conducting a systematic accuracy--efficiency analysis to achieve effective approximation without significantly compromising detection performance.
    
    \item We provide an extensive evaluation of the proposed methodology for visual novelty detection under different adverse scenarios including limited training samples, severe class imbalance, long-tailed distributions, {\em etc.}, and contrast its performance against state-of-the-art methods.
\end{itemize}

A preliminary version of the current study appeared in \cite{adsaei}. This paper extends the conference version in \cite{adsaei} in multiple aspects,  including: (i) extending the experimental evaluation on additional and challenging long-tailed recognition datasets, namely CIFAR-10-LT and CIFAR-100-LT \cite{krizhevsky2009learning,kim2020m2m}, and ImageNet-LT \cite{liu2019oltr}, which were not considered in the initial version; (ii) a systematic study of kernel approximation methods, including Nystr\"om, RFF, QMC-RFF, ORF, SORF, Fastfood, and RPCholesky, along with an accuracy--efficiency trade-off analysis to render the proposed approach a practical proposition on large datasets; (iii) a clarified description and extended convergence analysis of the proposed two-block optimization scheme; (iv) expanded comparisons against more recent approaches in different evaluation settings; and (v) additional analyzes covering joint versus fixed representation learning and statistical comparisons of different methods, in addition to rewriting the text for enhanced clarity.

\section{Related Work}
\label{sec:related}
The current study sits at the intersection of boundary-based novelty detection, deep representation learning, long-tailed recognition, and scalable kernel approaches. We briefly review these directions with an emphasis on explicit decision geometry learning, joint representation--boundary training, and evaluation under limited supervision, open-set uncertainty, and long-tailed distributions.

\subsection{Boundary-based and deep data-description methods}

Support Vector Data Description (SVDD)~\cite{tax2004support} and one-class SVMs form a classical foundation for novelty and anomaly detection by learning compact decision regions in a feature space. Their kernelized formulations provide nonlinear decision boundaries, while extensions modify the boundary geometry, slack penalty, or kernel combination strategy to improve robustness and flexibility. The $\ell_p$-norm SVDD formulation \cite{RAHIMZADEHARASHLOO2022108930} generalizes the conventional linear slack penalty, while the large-margin multiple-kernel $\ell_p$-SVDD method~\cite{arashloo2024large} explicitly maximizes a margin between normal and anomalous samples and solves the resulting convex dual efficiently. This method is the closest fixed-feature predecessor to our work: it provides a strong margin-aware kernel boundary, but does not provide a feature learning mechanism. Related image-based anomaly-detection studies also highlight the relevance of principled one-class objective functions, stable optimization, and computational efficiency, including $\ell_p$-norm multiple-kernel Fisher null learning~\cite{arashloo2023tipmkl}, adversarially learned one-class novelty detection with pseudo anomalies~\cite{zaheer2022pseudo}, and fast cascade patch retrieval for anomaly detection and localization~\cite{li2024target}.

Deep boundary-based methods aim to combine the geometric intuition of SVDD with the representation power of neural networks \cite{ruff2018deep,Ruff2020Deep,goyal2020drocc}. Recent methods of the SVDD family further address issues such as hypersphere collapse, insufficient feature structure, and multi-modal normal distributions \cite{hojjati2023dasvdd,lagrassa2020ocmst,xing2024dmsvdd}. Very recent maximum-margin deep anomaly-detection formulations, such as IMD-AD~\cite{yang2026imd}, also confirm the relevance of margin-aware decision geometry learning. However, unlike approaches that replace the explicit SVDD boundary with a simplified hypersphere, reconstruction, graph-based, or multi-sphere
objective, the proposed DLM-SVDD approach retains the original nonlinear kernel large-margin $\ell_p$-SVDD dual formulation during the boundary-fitting step. The CNN representation is subsequently updated using a loss function induced by the margin constraints of this explicitly recovered boundary.

\subsection{Representation learning and transfer-based anomaly detection}

A complementary line of anomaly detection work focuses on representation quality than the exact form of the decision boundary. As an example, PANDA~\cite{reiss2021panda} showed that adapting ImageNet-pretrained features can yield strong anomaly-detection performance when overfitting to the nominal class is controlled. Mean-Shifted Contrastive learning (MSC)~\cite{reiss2023msc} further improves pretrained encoders by modifying the contrastive objective to avoid feature collapse and produce more discriminative anomaly scores. Self-supervised and contrastive novelty detectors follow a similar representation-first philosophy. As an instance, CSI~\cite{tack2020csi} learns discriminative embeddings through contrastive learning on distributionally shifted views while Outlier Exposure~\cite{hendrycks2019oe} improves robustness using auxiliary outliers and UNODE~\cite{mirzaei2024unode} pursues universal novelty detection through adaptive contrastive learning whereas FIRM~\cite{lunardi2025firm} focuses on learning in-distribution representations for anomaly detection using contrastive objectives. Although these methods represent strong empirical baselines, they usually rely on post-hoc distance, density, or classifier-based scoring rather than optimizing the representation to satisfy an explicit large-margin data-description boundary.
Large pretrained vision-language models have also become influential in zero-shot and class-generalizable anomaly detection \cite{zhou2024anomalyclip,ma2025aaclip}. These approaches demonstrate the strength of large-scale pretraining and are complementary to DLM-SVDD which focuses on task-specific joint learning of a CNN representation and an explicit large-margin $\ell_p$-SVDD boundary.
\subsection{Open-set, OOD, and long-tailed evaluation protocols}

The novelty detection problem is closely related to open-set recognition and out-of-distribution (OOD) detection where models must reject samples that deviate from the target distribution. The OOD detection literature has explored uncertainty-based, energy-based, feature-distance-based, and test-time adaptation strategies \cite{isaacmedina2025feverood,chen2025debo,yang2025oodd,zhang2026adaodd}. These methods are typically evaluated under multiclass in-distribution (ID) or OOD protocols rather than one-vs-rest novelty detection, but they reinforce the importance of reliable thresholding and robust separation between seen and unseen data.

Long-tailed recognition introduces another practical difficulty where class frequencies are highly imbalanced, and tail classes may be confused with unknown or anomalous samples. Classical long-tailed methods address this issue through re-weighting, decoupled training, calibration, or specialized architectures~\cite{kang2020crt,zhong2021mislass,cui2021reslt}, while recent methods further modify classifier or feature-learning geometry. Some examples of the methods in this group include GLMC+SEL~\cite{jian2025sel}, which uses supervised exploratory learning, and BCE3S~\cite{fan2026bce3s}, which combines binary cross-entropy based joint, contrastive, and classifier-uniform learning. Long-tailed OOD methods are also relevant because imbalance can distort classification scores and cause in-distribution samples near the tail to be mistaken as outliers. In this context, COCL~\cite{miao2024cocl} uses calibrated outlier-class learning, while EAT~\cite{wei2024eat}, ImOOD~\cite{liu2024imood}, and AdaptOD~\cite{miao2024adaptod} study head-biased OOD scores, tail/OOD confusion, and outlier distribution adaptation.

\subsection{Kernel approximation and scalable boundary optimization}

Kernel methods provide expressive nonlinear decision functions, but their scalability is limited by kernel matrix construction, repeated kernel-vector operations, and memory constraints. As such, their utility in very large-scale datasets has been limited. In this sense, scalability has long been a central issue for kernel methods. Nystr\"om approximations~\cite{williams2000nystrom} reduce the effective rank of the kernel matrix by using landmark samples, while Random Fourier Features~\cite{rahimi2007random} approximate shift-invariant kernels through explicit randomized feature maps. Structured variants such as Orthogonal Random Features (ORF) and Structured Orthogonal Random Features (SORF)~\cite{yu2016orf}, as well as Fastfood~\cite{le2013fastfood}, reduce computational cost by exploiting orthogonality or fast Hadamard transforms. QMC-RFF~\cite{avron2016qmc} improves standard RFF through low-discrepancy frequency sampling, while Randomly Pivoted Cholesky (RPCholesky)~\cite{chen2025randomly} provides a data-dependent low-rank factorization of the kernel matrix. In this work, we study how these approximation methods enable efficient boundary updates during optimization—reducing kernel operations complexities for large datasets without sacrificing the large-margin objective.

\subsection{Positioning of the proposed approach}
The most direct predecessor to our work is the fixed-feature large-margin $\ell_p$-SVDD framework of~\cite{arashloo2024large}. Deep data-description methods improve representation learning or reduce collapse, but generally rely on simplified hypersphere objectives, reconstruction terms, or multi-sphere surrogates rather than the large-margin $\ell_p$-SVDD dual objective function. Representation-first and foundation-model methods achieve strong empirical performance or transferability, but typically separate feature adaptation from an explicit margin-aware data-description boundary. Long-tailed OOD methods address imbalance and unknown detection mainly through classifier calibration, energy regularization, or outlier distribution adaptation. In contrast, the proposed DLM-SVDD approach adapts the CNN representation based on an explicit nonlinear large-margin $\ell_p$-SVDD boundary, while incorporating scalable kernel approximation strategies for effective and efficient optimization.

\section{Methodology}
\label{sec:method}

\subsection{Problem setup and notation}
We consider supervised anomaly detection in the one-vs-rest setting under both conventional and long-tailed training distributions. For each target class, samples belonging to that class are treated as \emph{normal} and all remaining classes are treated as \emph{anomalous}. Let
\(
I_N
\)
and
\(
I_A
\)
denote the index sets of normal and anomalous training samples, respectively. The composition of the two sets depends on the training protocol: standard one-vs-rest experiments use a prescribed negative-to-positive training ratio, whereas long-tailed tasks inherit their sample frequencies from the corresponding long-tailed distribution. Given an input image
\(
\mathbf{x}\in\mathbb{R}^{H\times W\times C}
\),
a convolutional backbone with trainable parameters
\(
\omega
\)
produces a feature vector
\(
\phi_\omega(\mathbf{x})\in\mathbb{R}^d
\).
Our goal is to jointly learn:
(i) a feature representation in which normal and anomalous samples are well separated, and (ii) a large-margin SVDD boundary that preserves the geometric structure of the large-margin $\ell_p$-SVDD formulation of \cite{arashloo2024large}. To this end, we propose an alternating optimization strategy that iteratively updates the SVDD boundary given the most recent representations and then updates the CNN parameters under a fixed boundary condition.

\subsection{Large-margin $\ell_p$-SVDD on deep representations}
Let
\(
z_i=\phi_\omega(\mathbf{x}_i)
\)
denote the deep representation of sample
\(
\mathbf{x}_i
\) obtained from the backbone CNN.
For fixed CNN parameters
\(
\omega
\),
the large-margin $\ell_p$-SVDD objective can be written as
\begin{equation}
\label{eq:primal_objective_journal}
\begin{aligned}
\min_{r,\mathcal{C},\boldsymbol{\epsilon},\rho}\quad
& r^2-\nu\rho^2
+ c_1\sum_{i\in I_N}\epsilon_i^p
+ c_2\sum_{l\in I_A}\epsilon_l^p \\
\text{s.t.}\quad
& \|\Phi(z_i)-\mathcal{C}\|_{\mathcal{H}_k}^2
\le r^2-\rho^2+\epsilon_i, \quad \forall i\in I_N, \\
& \|\Phi(z_l)-\mathcal{C}\|_{\mathcal{H}_k}^2
\ge r^2+\rho^2-\epsilon_l, \quad \forall l\in I_A, \\
& \epsilon_i,\epsilon_l\ge 0,
\end{aligned}
\end{equation}
where
\(
\Phi(\cdot)
\)
denotes the implicit feature map associated with kernel
\(
k(\cdot,\cdot)
\),
\(
\mathcal{H}_k
\)
is the corresponding RKHS,
\(
\mathcal{C}
\)
is the center of the hypersphere,
\(
r
\)
is its radius, and
\(
\rho
\)
is the margin parameter. The coefficients
\(
c_1
\),
\(
c_2
\),
\(
\nu
\),
and
\(
p
\)
control the trade-off between compactness, margin maximization, and slack penalization. As compared with the classical SVDD, the objective in Eq. \eqref{eq:primal_objective_journal} introduces two important enhancements. First, it explicitly maximizes a margin between the normal and anomalous classes through the margin term
\(
\nu\rho^2
\).
Second, it applies an $\ell_p$-norm penalty on the slack variables, which allows non-linear penalization of boundary violation errors and yields a more flexible treatment of the errors.

\subsection{Alternating optimization}
\label{subsec:alternating}
\begin{figure*}[!t]
    \centering
    \includegraphics[width=1\textwidth]{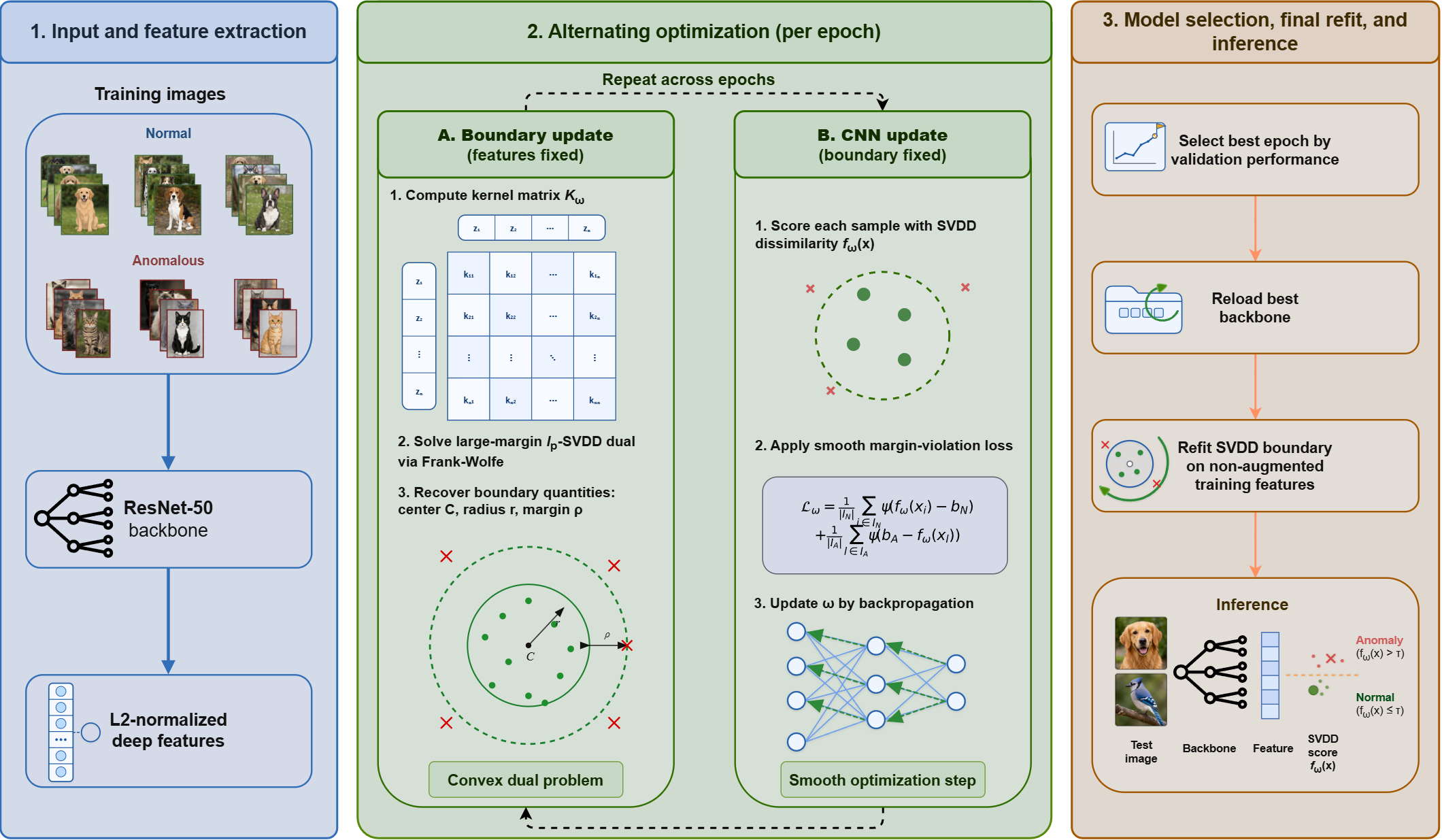}
    \caption{Overview of the proposed DLM-SVDD alternating optimization framework. At each epoch, the CNN backbone extracts deep features from all training samples.
    The $\alpha$-step constructs the RBF kernel matrix---using an exact or approximate backend---and solves the large-margin $\ell_p$-SVDD dual via Frank--Wolfe to recover the boundary parameters $(\boldsymbol{\alpha}^*,\mathcal{C}^*,r^*,\rho^*)$.
    The $\omega$-step fixes this boundary and updates the CNN parameters via backpropagation through the margin-violation loss $\mathcal{L}_\omega$. Validation scores are used for protocol-specific model selection:
    AUROC for the standard one-vs-rest benchmarks and threshold-optimized     balanced accuracy for the long-tailed experiments. After training, the     selected backbone is reloaded, the SVDD boundary is refitted from scratch using non-augmented training features, and test performance is computed from the resulting anomaly scores.}
    \label{fig:framework}
\end{figure*}

A direct and joint optimization of Eq. \eqref{eq:primal_objective_journal} with respect to both the boundary variables and CNN parameters is challenging as the SVDD boundary depends nonlinearly on the deep representation and the CNN update step is nonconvex. We therefore propose a two-block alternating strategy that is repeated over training epochs, as summarized in Fig.~\ref{fig:framework}.

\subsubsection{Boundary update under fixed representations}
At the beginning of each epoch, the CNN backbone is used to extract features from the penultimate layer of the CNN backbone. Given these features, the large-margin $\ell_p$-SVDD problem reduces to the kernel-space optimization of \cite{arashloo2024large}. Using
\(
\mathbf{t}\in\{\pm1\}^n
\)
to denote the labels, with
\(
t_i=+1
\)
for normal samples and
\(
t_i=-1
\)
for anomalous observations, and using
\(
\mathbf{K}_\omega
\)
to denote the kernel matrix computed for the current deep features, the dual problem takes the form:
\begin{align}
\nonumber
\min_{\boldsymbol{\alpha}}\quad
& \bar{c}_1
\|\boldsymbol{\alpha}\odot(\mathbf{1}+\mathbf{t})\|_q^q
+
\bar{c}_2
\|\boldsymbol{\alpha}\odot(\mathbf{1}-\mathbf{t})\|_q^q \\
&\hspace{2em}
+
(\boldsymbol{\alpha}\odot\mathbf{t})^\top
\mathbf{K}_\omega
(\boldsymbol{\alpha}\odot\mathbf{t})
\label{eq:dual_journal}\\
\text{s.t.}\quad
& \boldsymbol{\alpha}\ge 0,\quad
\mathbf{1}^\top\boldsymbol{\alpha}=\nu,\quad
\mathbf{t}^\top\boldsymbol{\alpha}=1.
\nonumber
\end{align}
Here,
\(
q=\frac{p}{p-1}
\)
is the conjugate exponent and
\(
\bar{c}_1=
\frac{1}{2}(c_1p)^{-\frac{1}{p-1}}
\left(1-\frac{1}{p}\right),
\qquad
\bar{c}_2=
\frac{1}{2}(c_2p)^{-\frac{1}{p-1}}
\left(1-\frac{1}{p}\right),
\) and \(
\boldsymbol{\alpha}
\) represents dual variables. Following~\cite{arashloo2024large}, we solve the problem in Eq.~\eqref{eq:dual_journal} using the Frank--Wolfe algorithm. Let \(F_\omega(\boldsymbol{\alpha})\) denote the objective in Eq.~\eqref{eq:dual_journal}, and let
\(
\mathcal{D}
=
\{\boldsymbol{\alpha}\geq 0:
\mathbf{1}^{\top}\boldsymbol{\alpha}=\nu,\,
\mathbf{t}^{\top}\boldsymbol{\alpha}=1\}
\)
denote its feasible set. At iteration \(\tau\), the Frank--Wolfe updates are
\begin{equation}
\label{eq:fw_updates}
\begin{aligned}
\mathbf{s}^{(\tau)}
&=
\arg\min_{\mathbf{s}\in\mathcal{D}}
\left\langle
\mathbf{s},
\nabla F_\omega\!\left(\boldsymbol{\alpha}^{(\tau)}\right)
\right\rangle,\\
\boldsymbol{\alpha}^{(\tau+1)}
&=
\left(1-\lambda_\tau\right)\boldsymbol{\alpha}^{(\tau)}
+
\lambda_\tau\mathbf{s}^{(\tau)},
\qquad
\lambda_\tau=\frac{2}{\tau+2}.
\end{aligned}
\end{equation}
For the simplex constraints in Eq.~\eqref{eq:dual_journal}, the linear subproblem has a closed-form solution ~\cite{arashloo2024large}. In our implementation, this step is performed at every epoch using the current deep features. Hence, the boundary is repeatedly re-estimated so that it remains consistent with the evolving representation.

\subsubsection{CNN update given the boundary}
Once the boundary parameters are derived for the current epoch, they are kept fixed while the backbone CNN parameters
\(
\omega
\)
are updated using a gradient descent scheme. The goal of this step is to adapt the representation so that normal samples move toward the interior of the learned boundary and anomalous samples away from it. Let
\(
f_\omega(\mathbf{x})
\)
denote the SVDD dissimilarity score induced by the current boundary solution defined as the squared distance between the representation of \(\mathbf{x}\) and the learned SVDD center in the reproducing kernel Hilbert space:
\begin{equation}
\label{eq:svdd_dissimilarity}
f_\omega(\mathbf{x})
=
\left\|
\Phi\!\left(\phi_\omega(\mathbf{x})\right)
-
\mathcal{C}
\right\|_{\mathcal{H}_k}^{2}.
\end{equation}
Smaller values of \(f_\omega(\mathbf{x})\) indicate that the sample is more compatible with the learned normal region. Using the radius and margin values obtained during the boundary-update step, we define the inner normal boundary and the outer anomalous boundary as
\begin{equation}
\label{eq:boundaries}
b_N=r^2-\rho^2,
\qquad
b_A=r^2+\rho^2.
\end{equation}

For a fixed boundary, the slack variables induced by the primal constraints are
\(
\epsilon_i(\omega)=[f_\omega(\mathbf{x}_i)-b_N]_{+}
\)
for \(i\in I_N\), and
\(
\epsilon_l(\omega)=[b_A-f_\omega(\mathbf{x}_l)]_{+}
\)
for \(l\in I_A\), where
\(
[u]_{+}=\max(0,u)
\).
Therefore, the CNN-update loss directly inherited from the
large-margin \(\ell_p\)-SVDD primal objective would be
\begin{equation}
\label{eq:cnn_powered_hinge_loss}
\mathcal{L}^{p\text{-hinge}}_{\omega}
=
\frac{1}{|I_N|}
\sum_{i\in I_N}
\left[
f_\omega(\mathbf{x}_i)-b_N
\right]_{+}^{p}
+
\frac{1}{|I_A|}
\sum_{l\in I_A}
\left[
b_A-f_\omega(\mathbf{x}_l)
\right]_{+}^{p}.
\end{equation}

For the CNN update step, in practice, for a more stable optimization, we employ a softplus function, \(\psi(u)=\log(1+e^u)\), as a smooth approximation to the hinge loss in Eq.~\eqref{eq:cnn_powered_hinge_loss}:
\begin{equation}
\label{eq:cnn_softplus_loss}
\mathcal{L}_{\omega}
=
\frac{1}{|I_N|}
\sum_{i\in I_N}
\psi^p\!\left(f_\omega(\mathbf{x}_i)-b_N\right)
+
\frac{1}{|I_A|}
\sum_{l\in I_A}
\psi^p\!\left(b_A-f_\omega(\mathbf{x}_l)\right).
\end{equation}

The difference between the two losses can be understood from their gradients. For \(p>1\), the powered hinge satisfies
\(
\partial [u]_{+}^{p}/\partial u
=
p[u]_{+}^{p-1}\mathbb{I}(u>0)
\),
whereas the softplus derivative is the bounded sigmoid function
\(
\psi'(u)=\sigma(u)=1/(1+\exp(-u))
\).
The powered hinge therefore assigns very small gradients to mild violations but potentially large gradients to severe ones, allowing a small number of highly violating or noisy samples to dominate the CNN update. The softplus surrogate instead provides a meaningful gradient near the boundary while limiting the influence of extreme violations, resulting in better-conditioned updates across alternating epochs.

To express the gradient-based CNN update, let
\(
v_i^{N}=f_\omega(\mathbf{x}_i)-b_N
\)
and
\(
v_l^{A}=b_A-f_\omega(\mathbf{x}_l)
\).
The gradient of Eq.~\eqref{eq:cnn_softplus_loss} with respect to the CNN parameters is
\begin{equation}
\label{eq:cnn_loss_gradient}
\begin{aligned}
\nabla_{\omega}\mathcal{L}_{\omega}
&={}
\frac{1}{|I_N|}
\sum_{i\in I_N}
\sigma\!\left(v_i^{N}\right)
\nabla_{\omega}f_\omega(\mathbf{x}_i)
\\
&-
\frac{1}{|I_A|}
\sum_{l\in I_A}
\sigma\!\left(v_l^{A}\right)
\nabla_{\omega}f_\omega(\mathbf{x}_l),
\end{aligned}
\end{equation}
\noindent where the first term decreases the dissimilarity scores of normal samples that violate the inner boundary, while the second increases the scores of anomalous samples that violate the outer boundary. The resulting gradient is used to update \(\omega\) using the Adam optimizer.

\subsection{Decision function}
The learned RKHS center is
\(
\mathcal{C}=\sum_{i=1}^{n}\alpha_i t_i \Phi(z_i)
\).
Let
\(
\mathbf{a}=\boldsymbol{\alpha}\odot\mathbf{t}
\).
For a test image
\(
\mathbf{x}
\)
with feature
\(
z=\phi_\omega(\mathbf{x})
\),
let
\(
\mathbf{k}(z)=[k(z,z_1),\ldots,k(z,z_n)]^\top
\). In this study, we use the widely adopted Gaussian (RBF) kernel function:
\begin{equation}
k(z_i,z_j)=\exp\!\left(-\gamma\|z_i-z_j\|_2^2\right).
\end{equation}
Since the RBF kernel satisfies
\(
k(z,z)=1
\),
the squared distance from the center is given as
\begin{equation}
\label{eq:decision_score}
f_\omega(\mathbf{x})
=
1
-
2\mathbf{k}(z)^\top\mathbf{a}
+
\mathbf{a}^\top\mathbf{K}_\omega\mathbf{a}.
\end{equation}
The score is interpreted as a dissimilarity measure: smaller values indicate stronger membership in the normal class, while larger values indicate anomalous behavior.
For recovering the radius and margin, let
\(
S_N=\{i\in I_N:\alpha_i>0\}
\)
and
\(
S_A=\{l\in I_A:\alpha_l>0\}
\)
denote the normal and anomalous support sets, with
\(
n'_N=|S_N|
\)
and
\(
n'_A=|S_A|
\).
The corresponding slack values are
\(
\epsilon_i=(\alpha_i/(c_1p))^{1/(p-1)}
\)
for
\(
i\in I_N
\)
and
\(
\epsilon_l=(\alpha_l/(c_2p))^{1/(p-1)}
\)
for
\(
l\in I_A
\).
Following the complementary slackness conditions, let us define
\begin{align}
A_N
&=
\frac{1}{n'_N}
\sum_{i\in S_N}
\big(f_\omega(\mathbf{x}_i)-\epsilon_i\big),
&
A_A
&=
\frac{1}{n'_A}
\sum_{l\in S_A}
\big(f_\omega(\mathbf{x}_l)+\epsilon_l\big).
\end{align}
The squared radius and total margin are then recovered as
\begin{equation}
\label{eq:radius_margin_recovery}
r^2=\frac{1}{2}(A_N+A_A),
\qquad
2\rho^2=A_A-A_N.
\end{equation}

\subsection{Kernel approximation strategies for improved scalability}
\label{subsec:approx}

Exact RBF kernel computations incur $\mathcal{O}(N^2d)$ time and $\mathcal{O}(BN)$ working memory (row-block size $B$) complexity for the initial kernel--vector product. Although subsequent Frank--Wolfe iterations cost only $\mathcal{O}(Nd)$ with on-the-fly column computations, the initial quadratic step remains a bottleneck for large $N$. We therefore use low-rank backends that construct $\widetilde{\mathbf{Z}}\in\mathbb{R}^{N\times m}$, where $m\ll N$ is the approximation budget: the number of Nystr\"om landmarks, spectral features, or RPCholesky pivots. With
$\mathbf{K}_\omega\approx
\widetilde{\mathbf{Z}}\widetilde{\mathbf{Z}}^\top$, for any
$\mathbf{v}\in\mathbb{R}^N$,
\begin{align}
    \mathbf{K}_\omega \mathbf{v} \approx \widetilde{\mathbf{Z}}\left(\widetilde{\mathbf{Z}}^\top \mathbf{v}\right),
    \quad
    \mathbf{K}_\omega[:,k] \approx \widetilde{\mathbf{Z}}\widetilde{\mathbf{Z}}_{k,:}^{\top},
\end{align}
are both computable in $\mathcal{O}(Nm)$ time, where $\widetilde{\mathbf{Z}}_{k,:}$ is the $k$-th row of $\widetilde{\mathbf{Z}}$.

\noindent\textbf{Nystr\"om approximation.} Given $m$ landmarks $\mathcal{U}$ sampled uniformly without replacement, let $\mathbf{C}=\mathbf{K}_{X,\mathcal{U}}\in\mathbb{R}^{N\times m}$ and $\mathbf{W}=\mathbf{K}_{\mathcal{U},\mathcal{U}}=\mathbf{U}\boldsymbol{\Lambda}\mathbf{U}^\top$. We set \(
\widetilde{\mathbf{Z}}=\mathbf{C}\mathbf{U}(\boldsymbol{\Lambda}+\delta\mathbf{I})^{-1/2}\) with small stabilizer $\delta\ge 0$, yielding $\mathbf{K}_\omega\approx\mathbf{C}\mathbf{W}^{\dagger}\mathbf{C}^\top\approx\widetilde{\mathbf{Z}}\widetilde{\mathbf{Z}}^\top$.

\noindent\textbf{Spectral (Fourier) methods.} By Bochner's theorem, the RBF kernel admits the randomized feature map
\begin{equation}
\label{eq:cosine_map}
\tilde z(x)=\sqrt{\tfrac{2}{m}}\left[\cos(\boldsymbol{\omega}_1^\top x+b_1),\ldots,\cos(\boldsymbol{\omega}_m^\top x+b_m)\right]^\top,
\end{equation}
where the phase offsets are drawn as $b_j\sim\mathrm{Unif}[0,2\pi]$ for $j=1,\ldots,m$. This map is shared by all approximation methods below; they only differ in how the frequency vectors $\boldsymbol{\omega}_j$ (rows of $\boldsymbol{\Omega}$) are generated, with $\widetilde{\mathbf{Z}}$ obtained by stacking $\tilde z(x_i)^\top$ row-wise for the full training set.
\begin{itemize}
    \item \emph{RFF}: $\boldsymbol{\omega}_j\sim\mathcal{N}(\mathbf{0},2\gamma\mathbf{I}_d)$, i.i.d.
    \item \emph{QMC-RFF}: $\boldsymbol{\omega}_j=\sqrt{2\gamma}\, \Phi_{\mathrm{G}}^{-1}(\mathbf{u}_j)$, where $\Phi_{\mathrm{G}}^{-1}$ is the componentwise standard-normal quantile and $\mathbf{u}_j\in(0,1)^d$ is a scrambled Sobol point used to reduce variance.
    \item \emph{ORF}: $\boldsymbol{\Omega}_{\mathrm{ORF}}=\sqrt{2\gamma}\,\mathbf{S}\mathbf{Q}$ per $d\times d$ block, with random orthogonal $\mathbf{Q}$ and $\mathbf{S}=\mathrm{diag}(s_1,\ldots,s_d)$, $s_i\sim\chi_d$; blocks are concatenated for $m>d$.
    \item \emph{SORF}: $\boldsymbol{\Omega}_{\mathrm{SORF}}=\sqrt{2\gamma d}\,\mathbf{H}\mathbf{D}_1\mathbf{H}\mathbf{D}_2\mathbf{H}\mathbf{D}_3$, with Walsh--Hadamard $\mathbf{H}$ and random diagonal sign matrices $\mathbf{D}_1,\mathbf{D}_2,\mathbf{D}_3$, reducing per-sample cost from $O(dm)$ to $O(m\log d)$.
    \item \emph{Fastfood}: $\boldsymbol{\Omega}_{\mathrm{FF}}=\sqrt{2\gamma/d}\,\mathbf{S}\mathbf{H}\mathbf{G}\boldsymbol{\Pi}\mathbf{H}\mathbf{B}$, with sign matrix $\mathbf{B}$, permutation $\boldsymbol{\Pi}$, diagonal Gaussian $\mathbf{G}$, and scaling $\mathbf{S}$, also achieving $O(m\log d)$ generation cost via structured transforms rather than a dense Gaussian matrix.
\end{itemize}
SORF and Fastfood thus reduce feature construction to $O(Nm\log d)$, versus $O(Ndm)$ for RFF/QMC-RFF/ORF.

\noindent\textbf{Randomly Pivoted Cholesky (RPCholesky).} RPCholesky selects pivots sequentially with probability proportional to the current residual diagonal of $\mathbf{K}_\omega$, building a rank-$m$ factor $\mathbf{L}\in\mathbb{R}^{N\times m}$ so that $\widetilde{\mathbf{Z}}=\mathbf{L}$ and $k(x,y)\approx\tilde z(x)^\top\tilde z(y)$.

\subsection{Convergence analysis}
\label{subsec:convergence}

The proposed optimization approach alternates between a convex boundary update and a gradient-based representation update. We adopt two standard conditions: (A1) Bounded trajectory --- the iterates $\{\omega^{(s)}\}$ remain in a compact set $\Omega\subset\mathbb{R}^{d_\omega}$, promoted in practice by the weight decay and gradient-norm clipping (Section~\ref{subsec:training_config}) --- and (A2) Local smoothness --- for every sample $x$, $\omega\mapsto f_\omega(x)$ is differentiable with $L_f$-Lipschitz gradient on $\Omega$. Since the training set is finite and \(\Omega\) is compact, (A1)--(A2) and continuity further imply
\(
G=\max_{\omega\in\Omega,\,x\in\mathcal X_{\mathrm{tr}}}
\|\nabla_\omega f_\omega(x)\|_2<\infty
\).

\textbf{Boundary update.}
For fixed \(\omega\), the \(\alpha\)-step minimizes
\(F_\omega(\boldsymbol\alpha)\) in Eq.~\eqref{eq:dual_journal} over
\(
\mathcal D=\{\boldsymbol\alpha\geq\mathbf 0:
\mathbf 1^\top\boldsymbol\alpha=\nu,\,\mathbf t^\top\boldsymbol\alpha=1\}
\),
a compact polytope for \(\nu\geq1\). For \(p>1\), \(F_\omega\) is convex, while for the common setting \(p=2\), used in the expanded experiments and selected in most standard one-vs-rest tasks, \(q=2\) and
\(F_\omega\) is a convex quadratic function in \(\boldsymbol\alpha\) with a constant Hessian. Compactness of \(\mathcal D\) therefore gives a finite curvature constant
\(
C_F\leq\operatorname{diam}(\mathcal D)^2\|\nabla^2F_\omega\|_{\mathrm{op}}
\),
and the Frank--Wolfe iterates converge to the global minimizer of the fixed-boundary subproblem at the standard rate
\(
F_\omega(\boldsymbol\alpha^{(\tau)})-F_\omega^*\leq 2C_F/(\tau+2)
\)~\cite{jaggi2013revisiting}, where
\(F_\omega^*=\min_{\boldsymbol\alpha\in\mathcal D}F_\omega(\boldsymbol\alpha)\).
For any \(p>1\), the computable Frank--Wolfe gap \begin{equation}
\label{eq:fw_gap}
0\leq F_\omega(\boldsymbol\alpha)-F_\omega^*\leq
g_{\mathrm{FW}}(\boldsymbol\alpha;\omega):=
\max_{\mathbf s\in\mathcal D}
\langle\boldsymbol\alpha-\mathbf s,
\nabla F_\omega(\boldsymbol\alpha)\rangle
\end{equation}
is a valid suboptimality certificate and stopping criterion for the boundary solver~\cite{jaggi2013revisiting}.

\textbf{Network update.}
For fixed boundary parameters, the network step minimizes the softplus margin-violation objective \(\mathcal L_\omega\) in Eq.~\eqref{eq:cnn_softplus_loss}. Since the softplus function satisfies \(0\leq\psi'\leq1\) and \(\psi''\leq1/4\), each term \(\psi(f_\omega(x_i)-b_N)\) (similarly for the anomalous term) has gradient-Lipschitz constant at most \(G^2/4+L_f\) by the standard composition rule for smooth functions under (A1)--(A2); summing the two normalized class-wise averages gives $L_\omega\leq\tfrac{1}{2}G^2+2L_f<\infty$. A gradient step with \(0<\eta\leq1/L_\omega\) then satisfies the descent inequality~\cite{nesterov2018lectures}
\begin{equation}
\label{eq:cnn_descent}
\mathcal L_\omega(\omega^{(s+1)})\leq
\mathcal L_\omega(\omega^{(s)})-\tfrac{\eta}{2}
\|\nabla_\omega\mathcal L_\omega(\omega^{(s)})\|_2^2,
\end{equation}
and since \(\mathcal L_\omega\geq0\), summing Eq.~\eqref{eq:cnn_descent} over iterations shows the loss values converge and \(\|\nabla_\omega\mathcal L_\omega(\omega^{(s)})\|_2\to0\).

\textbf{Alternating behavior.}
Let \(\boldsymbol\alpha^{(t)}\) be the boundary obtained for \(\omega^{(t)}\) and \(\omega^{(t+1)}\) be the parameters of the subsequent update step, with tolerance sequences \(\varepsilon_t^\alpha,\varepsilon_t^\omega\to0\) such that
\(
g_{\mathrm{FW}}(\boldsymbol\alpha^{(t)};\omega^{(t)})\leq\varepsilon_t^\alpha
\)
and
\(
\|\nabla_\omega\mathcal L_\omega(\omega^{(t+1)};\boldsymbol\alpha^{(t)})\|_2\leq\varepsilon_t^\omega
\),
and assume asymptotic regularity,
\(\|\omega^{(t+1)}-\omega^{(t)}\|_2\to0\). By compactness of
\(\Omega\times\mathcal D\) under (A1) and continuity of
\(g_{\mathrm{FW}}\) and \(\nabla_\omega\mathcal L_\omega\) under (A2),
every accumulation point \((\omega^*,\boldsymbol\alpha^*)\) satisfies
\begin{equation}
\label{eq:block_stationarity}
\boldsymbol\alpha^*\in\arg\min_{\boldsymbol\alpha\in\mathcal D}F_{\omega^*}(\boldsymbol\alpha),
\qquad
\nabla_\omega\mathcal L_\omega(\omega^*;\boldsymbol\alpha^*)=\mathbf 0,
\end{equation}
{\em i.e.}, the boundary \textit{in the limit} is globally optimal for the corresponding representation while the representation is stationary for the induced margin-violation loss; as the network update block is nonconvex, the guarantee concerns stationarity rather than global optimality in \(\omega\). The finite-iteration behavior of the two update blocks is empirically studied in Section~\ref{subsec:convergence_curves}.

\section{Experiments}
\label{sec:experiments}

\subsection{Datasets, evaluation scenarios, and performance metrics.}
We evaluate the proposed approach on multiple widely used and standard anomaly detection benchmarks as well as more challenging long-tailed protocols. For the standard anomaly detection setting, one class is treated as normal and the remaining classes are treated as anomalous, yielding one binary anomaly detection task per class. Following the common practice from the literature and for a fair evaluation, we report the performance by averaging performances across all classes. The datasets used in our experiments are Fashion-MNIST~\cite{xiao2017fashion}, CIFAR-10 and CIFAR-100~\cite{krizhevsky2009learning}, and ImageNet-LT~\cite{liu2019oltr}. Following the long-tailed CIFAR construction protocol adopted in~\cite{kim2020m2m}, we construct CIFAR-10-LT and CIFAR-100-LT from the corresponding CIFAR training sets using imbalance ratios 
\(
\varrho\in\{10,50,100\}
\), which control the severity of the class-imbalanced. Following the standard long-tailed CIFAR training set construction protocol from the literature, the training set cardinality of class
\(
c\in\{0,\dots,C-1\}
\)
is
\begin{equation}
\label{eq:lt_formula}
n_c =
\mathrm{round}\!\left(
n_0\varrho^{-\frac{c}{C-1}}
\right),
\end{equation}
where
\(
C
\)
is the total number of classes and
\(
n_0
\)
is the training cardinality of the first, head class in the training set. The class frequencies decay exponentially from head class towards the tail classes, and the last class contains approximately
\(
1/\varrho
\)
as many training examples as the first class. Following the common practice in the literature and in order to enable a fair comparison, in the long-tailed setting, the resulting class-imbalanced training pool is converted into one-vs-rest anomaly detection tasks. For each task, we perform a stratified 80/20 split into training and validation sets. The final testing is performed on the full balanced test set of the corresponding dataset. 

In the standard one-vs-rest anomaly-detection benchmarks, the primary evaluation metric we use is the area under the receiver operating characteristic curve (AUROC). For this purpose, AUROC is computed separately for each normal class and then averaged across classes. For thresholded evaluation, the decision threshold is set on the validation set by maximizing balanced accuracy, as described in Section~\ref{subsec:impl}. For the long-tailed recognition experiments, we follow the widely used protocol in~\cite{arashloo2024large}: a separate one-vs-rest detector is trained for each class. The validation set is then used to select the decision threshold and test performance is reported using balanced accuracy computed as the average class-wise recall. In addition to the standard performance metrics, in order to summarize comparisons involving many baselines (with potentially missing entries) we use the Skillings--Mack test~\cite{skillings1981general}, a Friedman-type omnibus rank test capable of handling incomplete/missing performances for statistical ranking. The resulting statistical ranking significance values, $p$-values, are reported in the captions of Tables~\ref{tab:overall_results} and~\ref{tab:lt_results} where $p<0.05$ is interpreted as statistically significant evidence of global rank differences.

\subsection{Implementation details}
\label{subsec:impl}
\textbf{CNN backbone, feature normalization, and kernel choice.}
Our backbone CNN network is ResNet-50~\cite{he2016deep}. For Fashion-MNIST, CIFAR-10, CIFAR-100, CIFAR-10-LT, and CIFAR-100-LT, the backbone is initialized using supervised ImageNet-pretrained weights. For ImageNet-LT, we instead use a DINO-pretrained ResNet-50 initialization~\cite{caron2021emerging}. Since ImageNet-LT shares the ImageNet-1K label space, supervised ImageNet initialization would expose the backbone to class-label supervision from the same 1,000 categories before long-tailed training. We therefore use self-supervised DINO initialization to avoid direct class-label bias and supervision information leakage during training. In all cases, the final classification layer is removed, and the resulting feature vector is $\ell_2$-normalized before kernel computation. For the RBF kernel function, we adopt a fixed, once-estimated bandwidth $\gamma$, set based on the median squared pairwise Euclidean distances of the pretrained training features which is kept constant throughout training; this choice is justified against two alternative bandwidth strategies in the ablation study (Section~\ref{subsec:ablation}).

\textbf{Data preprocessing and augmentation.}
All images are normalized using dataset-specific channel statistics. For CIFAR-10 and CIFAR-100, we apply standard data augmentation consisting of random cropping with padding and random horizontal flipping during training. Validation and test images are used without augmentation. For Fashion-MNIST, only normalization is applied. As spatial augmentations such as random cropping on this dataset may remove informative object parts or introduce distortions, they are not used on this dataset.

\textbf{Training configuration.}\label{subsec:training_config}
The training stage proceeds for a fixed number of epochs using the alternating optimization procedure described in Section~\ref{sec:method}. At each epoch:
\begin{enumerate}
    \item the current CNN backbone yields features for the full training set;
    \item an exact/approximate RBF kernel matrix is constructed from these features;
    \item the large-margin $\ell_p$-SVDD dual is solved to recover the boundary parameters $(\boldsymbol{\alpha},\mathcal{C},r,\rho)$;
    \item the CNN is updated following a mini-batch approach using Adam optimizer on the smooth margin-violation loss in Eq.~\eqref{eq:cnn_softplus_loss};
    \item validation scores are used to select the best model and the corresponding decision threshold.
\end{enumerate}

For the standard one-vs-rest experiments, the potential values of the parameters are: \(p\in\{1.01,1.1,1.2,2,5,10,100\}\), \(\nu\in\{1,1.2,1.5,1.7,2,5\}\), and learning rates \(\eta\in\{10^{-k},5\times10^{-(k+1)}\}_{k=2}^{5}\). Based on our observations on validation data, \(p=2\) and \(\nu=1.2\) were selected for the majority of classes. As such, we adopt \(p=2\), \(\nu=1.2\), and \(\eta=10^{-4}\) as a common configuration for the the experiments to limit task-specific tuning and ensure consistent comparisons. We use weight decay \(10^{-4}\), batch size of \(128\), gradient clipping at \(1\), \(c_1=1/|I_N|\), \(c_2=1/|I_A|\), and fix the maximum number of Frank--Wolfe iterations to \(1000\). 

In our preliminary experiments, the softplus function used as a smooth approximation to the hinge loss led to smooth and effective optimization; however, setting \(p\) to $1$ in the softplus function led to even more stable optimization. We therefore use this choice in the softplus function in all reported experiments. Note that this choice affects only the weighting of violations during the CNN-update step. The parameter \(p\) and its corresponding nonlinear slack penalization remain unchanged in the exact large-margin \(\ell_p\)-SVDD boundary-update problem. For all experiments, \(20\%\) of the available training data is held out for validation using a stratified split. For the long-tailed experiments, we train for \(500\) epochs where the best epoch is selected based on balanced validation accuracy after threshold optimization.

\subsection{Anomaly detection}
We compare the proposed approach with both classical as well as recent anomaly detection approaches, including kernel-based SVDD-style models, deep boundary-based methods, and recent representation-learning approaches. In particular, we include the fixed-feature large-margin $\ell_p$-SVDD baseline of~\cite{arashloo2024large}, which is the most direct reference point for assessing the benefit of joint representation adaptation. Wherever applicable, we shall use the strongest single-kernel variant to ensure a fair comparison against our single-kernel deep model.

Table~\ref{tab:overall_results} summarizes the results of anomaly detection on the Fashion-MNIST, CIFAR-10, and CIFAR-100 datasets along with a statistical comparison and ranking of different approaches based on the Skillings--Mack test. To clarify the supervision levels of different methods, pure one-class methods that utilize no negative samples in the training stage are marked by $\star$, while methods using outlier exposure, synthetic outliers, contaminated-data assumptions, or auxiliary evidence are marked by $\diamond$. The negative-to-positive ratio, outlier exposure setting, or contamination setting is shown in front of each method whenever applicable. Unless otherwise stated, DLM-SVDD results are averaged over ten random seeds using matched data splits across variants. Since not all approaches report results on all datasets, average ranks are computed based on the reported entries only. Rows are grouped by supervision/protocol type rather than by rank, so that methods with different supervision settings may be compared and interpreted separately. For methods requiring additional clarification, CIFAR-100 results are marked by
$^{\dagger}$, contaminated or latent outlier exposure by $^{\ddagger}$, synthetic adaptive outlier exposure by $^{\S}$, and post-hoc contaminated-data adjustment with auxiliary evidence by $^{\P}$.

In the standard one-vs-rest anomaly-detection setting, the proposed DLM-SVDD approach achieves AUROCs of $0.977$, $0.982$, and $0.984$ on the CIFAR-10 dataset at ratios of 10\%, 50\%, and 75\%, respectively, and AUROCs of $0.962$, $0.971$, and $0.972$ on the CIFAR-100 dataset at the same ratios (Table~\ref{tab:overall_results}). The average-rank summary further clarifies the comparison across different methods: among the methods evaluated on all three standard benchmarks, the three DLM-SVDD variants obtain the best average ranks, namely 1.83, 3.00, and 5.00. These results match or exceed all pure one-class methods and all methods that do not use outlier exposure, synthetically generated outliers, contaminated-data assumptions, or auxiliary evidence. PANDA with outlier exposure achieves comparable AUROC rates on the CIFAR-10 and the CIFAR-100 datasets, but heavily relies on auxiliary outlier data, and thus, does not strictly follow similar evaluation protocols for a fair comparison. Under the same within-protocol anomaly-supervision setting, the proposed DLM-SVDD approach achieves the strongest overall performance and consistently improves over the fixed-feature large-margin $\ell_p$-SVDD baseline of~\cite{arashloo2024large} at all tested ratios on the CIFAR-10 and CIFAR-100 datasets. This confirms that the proposed joint representation--boundary learning approach possesses a practically meaningful merit over the fixed-feature predecessor.

\begin{table}[!t]
\centering
\footnotesize
\caption{Average test performances (in AUROC) on standard anomaly-detection benchmarks along with statistical average ranks. Rows are grouped by supervision/protocol type. Skillings--Mack: \(p_{\mathrm{SM}}\approx1.7\times10^{-4}\).}
\label{tab:overall_results}
\begin{adjustbox}{max width=1\columnwidth} 
\begin{tabular}{l c c c c}
\hline
\textbf{Method} & \textbf{F-MNIST} & \textbf{CIFAR-10} & \textbf{CIFAR-100} & \textbf{Rank} \\
\hline
\multicolumn{5}{l}{\emph{One-class, self-supervised, and no-OE baselines}} \\
Deep SVDD$\star$ \cite{ruff2018deep} & -- & 0.648 & -- & 28.00  \\
OCMST$\star$ \cite{lagrassa2020ocmst} & 0.873 & 0.748 & -- & 22.00 \\
DMSVDD-DRL$\star$ \cite{xing2024dmsvdd} & 0.946 & 0.712 & $0.687^{\dagger}$ & 20.00 \\
PANDA$\star$ (no OE) \cite{reiss2021panda} & 0.956 & 0.962 & 0.941 & 11.00 \\
MSC$\star$ (no OE) \cite{reiss2023msc} & -- & 0.972 & 0.964 & 6.25 \\
UNODE$\star$ \cite{mirzaei2024unode} & 0.934 & 0.969 & 0.936 & 12.00 \\
FIRM$\star$ \cite{lunardi2025firm} & 0.968 & 0.934 & $0.878^{\dagger}$ & 13.67 \\
The work $\star$ in \cite{10012327} & 0.937 & 0.925 & 0.882 & 16.67 \\
CSI \cite{tack2020csi} & -- & 0.943 & 0.896 & 15.00 \\
DROCC \cite{goyal2020drocc} & -- & 0.752 & -- & 25.00 \\
DROC-contrastive \cite{sohn2021learning} & 0.958 & 0.925 & 0.865 & 16.50 \\
ADMM-SRNet \cite{10124145} & -- & 0.954 & 0.916 & 13.50 \\
IMD-AD \cite{yang2026imd} & 0.965 & 0.834 & -- & 15.00 \\
\hline
\multicolumn{5}{l}{\emph{Methods with external or auxiliary anomaly evidence}} \\
Outlier Exposure$^{\diamond}$ \cite{hendrycks2019oe} & -- & 0.978 & 0.879 & 11.00 \\
PANDA (with OE)$^{\diamond}$ \cite{reiss2021panda} & 0.918 & 0.989 & 0.973 & 6.00 \\
LOE$^{\diamond}$ \cite{qiu2022loe}$^{\ddagger}$ & 0.929 & 0.949 & -- & 15.50 \\
UniCon-HA+OE$^{\diamond}$ \cite{wang2023uniconha} & -- & 0.969 & -- & 11.00 \\
FIRM+OE$^{\diamond}$ \cite{lunardi2025firm} & 0.962 & 0.976 & $0.947^{\dagger}$ & 7.50 \\
RODEO$^{\diamond}$ \cite{mirzaei2025rodeo}$^{\S}$ & 0.948 & 0.931 & 0.866 & 16.33\\
RD+EPHAD$^{\diamond}$ \cite{patra2025ephad}$^{\P}$ & 0.958 & 0.984 & -- & 5.50 \\
\hline
\multicolumn{5}{l}{\emph{Labeled-anomaly / ratio-based methods}} \\
Deep-SAD \cite{Ruff2020Deep} (ratio=10\%) & 0.882 & 0.860 & 0.891 & 18.00\\
Deep-SAD \cite{Ruff2020Deep} (ratio=75\%) & 0.981 & 0.925 & 0.887 & 13.00 \\
Single kernel \cite{arashloo2024large} (ratio=10\%) & -- & 0.958 & 0.947 & 11.25 \\
Single kernel \cite{arashloo2024large} (ratio=50\%) & -- & 0.969 & 0.953 & 9.00 \\
Single kernel \cite{arashloo2024large} (ratio=75\%) & -- & 0.972 & 0.958 & 7.25 \\
DLM-SVDD (ratio=10\%) & 0.976 & 0.977 & 0.962 & 5.00 \\
DLM-SVDD (ratio=50\%) & 0.987 & 0.982 & 0.971 & 3.00 \\
DLM-SVDD (ratio=75\%) & 0.988 & 0.984 & 0.972 & 1.83 \\
\hline
\end{tabular}\end{adjustbox}

\vspace{0.5mm}
\begin{minipage}{\columnwidth}
\scriptsize
\emph{Note:} $\star$ and $\diamond$ indicate supervision assumptions as defined in the text. $^{\dagger}$, $^{\ddagger}$, $^{\S}$, and $^{\P}$ denote the special reporting conditions described before the table. Average ranks are computed over reported entries only, with higher AUROC assigned better rank.
\end{minipage}
\end{table}


\subsection{Long-tailed recognition}
Table~\ref{tab:lt_results} reports the results under exponentially long-tailed training distributions. Following~\cite{arashloo2024large}, the proposed DLM-SVDD approach is applied to the long-tailed recognition problem by constructing a separate one-vs-rest detector model for each class and classifying each test sample. Performance is reported as balanced accuracy, computed as the average class-wise recall. ``Single-kernel $\ell_p$-SVDD'' denotes the fixed-feature baseline of~\cite{arashloo2024large}, while ``DLM-SVDD (fixed CNN)'' and ``DLM-SVDD (single ResNet-50, joint)'' denote our frozen-backbone variant and the proposed jointly trained model, respectively. Following the convention in the literature, the results for DLM-SVDD are averaged over ten seeds on the CIFAR-LT  datasets.

\begin{table*}[!t]
\centering
\footnotesize
\caption{Comparison with representative long-tailed recognition methods and SVDD-family variants on the CIFAR-10-LT, CIFAR-100-LT, and ImageNet-LT datasets in terms of balanced accuracy (\%). ``--'' denotes unavailable results. Skillings--Mack: \(p_{\mathrm{SM}}<10^{-6}\).}
\label{tab:lt_results}
\resizebox{\textwidth}{!}{%
\begin{tabular}{l ccc ccc cccc c}
\hline
\footnotesize
\textbf{Method} 
& \multicolumn{3}{c}{\textbf{CIFAR-10-LT}} 
& \multicolumn{3}{c}{\textbf{CIFAR-100-LT}} 
& \multicolumn{4}{c}{\textbf{ImageNet-LT}}
& \textbf{Avg. rank} \\
\cline{2-11}
& $\varrho=100$ & $\varrho=50$ & $\varrho=10$
& $\varrho=100$ & $\varrho=50$ & $\varrho=10$
& Many & Medium & Few & All
& $\downarrow$ \\
\hline
M2m \cite{kim2020m2m}
& 79.10 & -- & 87.50
& 43.50 & -- & 57.60
& -- & -- & -- & 43.70
& 13.80 \\

Ensemble \cite{li2022nested}
& 85.50 & 87.30 & --
& 54.20 & 58.20 & --
& -- & -- & -- & 60.50
& 5.80 \\

The work in \cite{zhao2023weightguided}
& 84.60 & 86.40 & --
& 55.10 & 58.70 & --
& 57.90 & 43.40 & 27.60 & 55.00
& 9.62 \\

MRA \cite{xiang2023mra}
& 82.30 & -- & 89.20
& 48.10 & -- & 62.20
& 53.00 & 40.80 & 27.30 & 43.60
& 13.12 \\

ResLT \cite{cui2021reslt}
& 82.40 & 85.10 & 89.70
& 49.70 & 54.50 & 63.70
& 63.60 & 55.70 & 38.90 & 56.10
& 10.45 \\

PaCo \cite{cui2021paco}
& -- & -- & --
& 52.00 & 56.00 & 64.20
& -- & -- & -- & 57.00
& 9.88 \\

BCL \cite{zhu2022bcl}
& 84.30 & 87.20 & 91.10
& 51.90 & 56.60 & 64.90
& 67.20 & 53.10 & 32.90 & 51.90
& 9.10 \\

DiffuLT \cite{shao2024diffult}
& 84.70 & 86.90 & 90.70
& 51.50 & 56.30 & 63.80
& 63.30 & 55.60 & 39.40 & 56.40
& 8.95 \\

LTRL \cite{zhao2024ltrl}
& -- & -- & --
& 52.00 & 56.80 & 64.70
& 69.00 & 56.00 & 38.40 & 58.40
& 7.14 \\

LA+Focal-SAM \cite{li2025focalsam}
& 82.90 & 85.50 & 90.50
& 50.70 & 54.50 & 63.80
& 63.90 & 52.20 & 34.40 & 54.30
& 10.70 \\

Multi-Granularity \cite{liu2025multigranularity}
& 87.50 & 88.16 & 92.37
& 54.02 & 58.13 & 65.86
& 71.50 & 56.80 & 41.50 & 60.40
& 5.00 \\

GLMC+SEL \cite{jian2025sel}
& 85.40 & 88.57 & 92.83
& 56.48 & 61.13 & 70.75
& 68.67 & 54.37 & 38.28 & 57.24
& 5.80 \\

BCE3S \cite{fan2026bce3s}
& 90.08 & 92.55 & 95.71
& 59.50 & 65.23 & 76.13
& 68.14 & 55.90 & 40.56 & 57.85
& 3.80 \\

Single-kernel $\ell_p$-SVDD \cite{arashloo2024large}
& 89.20 & 90.20 & 89.60
& 73.30 & 78.70 & 88.70
& 89.50 & 81.30 & 65.10 & 83.40
& 2.90 \\

DLM-SVDD (fixed CNN)
& 59.50 & 65.18 & 58.84
& 59.38 & 66.15 & 64.70
& 91.25 & 88.97 & 78.55 & 88.43
& 5.75 \\

DLM-SVDD (single ResNet-50, joint)
& 86.79 & 89.11 & 92.59
& 75.25 & 76.84 & 85.77
& 92.21 & 85.00 & 72.20 & 86.03
& 2.20 \\
\hline
\end{tabular}%
}
\end{table*}

As may be seen from Table~\ref{tab:lt_results}, the proposed jointly trained DLM-SVDD method obtains the best overall average rank, followed by the fixed-feature single-kernel $\ell_p$-SVDD baseline and BCE3S methods. Under long-tailed training distributions (Table~\ref{tab:lt_results}), the results highlight the importance of adapting the representation to the target imbalance regime, while also showing that the effect depends on the number of available positive samples. The jointly trained DLM-SVDD obtains the best average overall rank among all reported results and substantially improves over its fixed-CNN baseline on CIFAR-10-LT and CIFAR-100-LT. On the CIFAR-10-LT dataset, for instance, joint training improves the balanced accuracy from $59.50\%$, $65.18\%$, and $58.84\%$ to $86.79\%$, $89.11\%$, and $92.59\%$ at imbalance ratios $\varrho=100$, $50$, and $10$, respectively, with comparable gains observed on CIFAR-100-LT, emphasizing the merits of an end-to-end optimization of the large-margin decision boundary.

On the ImageNet-LT dataset, the trend is more nuanced: the jointly trained model improves the Many-shot split, whereas the fixed-CNN variant remains stronger on the Medium- and Few-shot splits. This suggests that when the positive training set is extremely small, updating the backbone can make the representation more sensitive to the limited available target samples, while frozen pretrained features may provide a more stable description for the tail classes. As the number of positive samples increases, this effect is reduced, and joint representation--boundary learning becomes more beneficial as expected. Overall, the proposed DLM-SVDD approach achieves the best average rank among different methods for the long-tailed recognition task, ahead of the fixed-feature single-kernel $\ell_p$-SVDD baseline and representative recent long-tailed recognition methods from the literature.

In this context, specialized long-tailed recognition methods such as BCE3S~\cite{fan2026bce3s} should be interpreted in light of their different objective and protocol. BCE3S is designed as a fully supervised multi-class recognition method and directly optimizes a class-wise classifier with joint feature--classifier learning, contrastive feature learning, and classifier uniform-separability learning. In contrast, DLM-SVDD follows a one-vs-rest anomaly-detection formulation, where each class is described by a separate large-margin SVDD boundary and decisions are made using validation-selected thresholds.


\subsection{Empirical optimization behavior}
\label{subsec:convergence_curves}

\begin{figure}[!t]
    \centering
    \subfloat[\scriptsize Relative Frank--Wolfe gap.]{
        \includegraphics[width=0.23\textwidth]
        {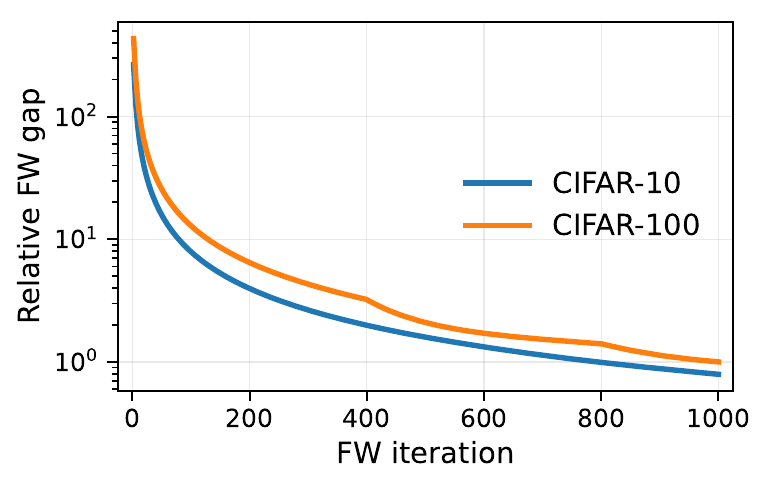}
    }
    \hfill
    \subfloat[\scriptsize CNN margin-violation loss.]{
        \includegraphics[width=0.23\textwidth]
        {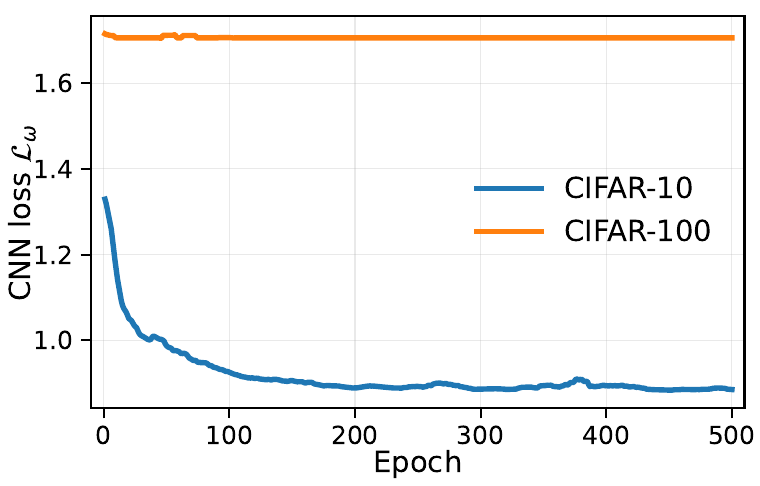}
    }
    \caption{Representative blockwise optimization behavior for DLM-SVDD on sample CIFAR-10 and CIFAR-100 classes with a negative-to-positive training ratio of \(10\%\): \textbf{(a)} relative Frank--Wolfe gap during a representative fixed-feature boundary update; \textbf{(b)} CNN margin-violation loss across outer training epochs. The CNN loss curves show an 11-epoch centered moving average.}
    \label{fig:convergence_curves}
\end{figure}

Fig.~\ref{fig:convergence_curves} illustrates the blockwise optimization behavior of DLM-SVDD on representative CIFAR-10 and CIFAR-100 classes. Panel~(a) shows the Frank--Wolfe gap defined as
\(
\widehat g_{\mathrm{FW}}^{(\tau)}
=
g_{\mathrm{FW}}^{(\tau)}/g_{\mathrm{FW}}^{(1)}
\)
during the fixed-feature boundary update, while panel~(b) shows the CNN margin-violation loss across outer training epochs. As may be observed from the figure, the Frank--Wolfe gap decreases substantially over the inner iterations, whereas the CNN loss decreases during the initial training stage and subsequently approaches a stable operating regime.

In particular, the behavior of the optimization in panel~(a) is consistent with the progress of the Frank--Wolfe solver toward the optimum of the fixed-feature boundary subproblem, supporting the analysis in Section~\ref{subsec:convergence}. Panel~(b) shows that the representation-update loss reaches a stable regime across outer epochs. Together, the two panels provide empirical evidence of stable optimization behavior in both update blocks within a limited number of iterations.

\subsection{Kernel approximation trade-offs}
\label{subsec:approx_results}

\begin{table}[!t]
\centering
\caption{Complexity of kernel approximation methods. $N$: samples, $d$: feature dimension, $m$: approximation rank or feature dimension.}
\label{tab:approx_complexity}
\footnotesize
\renewcommand{\arraystretch}{1.12}
\setlength{\tabcolsep}{0pt}
\begin{tabular*}{\columnwidth}{@{\extracolsep{\fill}}lccc@{}}
\hline
\textbf{Backend} & \textbf{Build} & \textbf{FW iter.} & \textbf{Memory} \\
\hline
Exact & $O(N^2d)$ & $O(Nd)$ & $O(Nd)$ / $O(N^2)$ \\
Nystr\"om & $O(Nmd+m^3)$ & $O(Nm)$ & $O(Nm+m^2)$ \\
RFF & $O(Ndm)$ & $O(Nm)$ & $O(Nm+dm)$ \\
QMC-RFF & $O(Ndm)$ & $O(Nm)$ & $O(Nm+dm)$ \\
ORF & $O(Ndm)$ & $O(Nm)$ & $O(Nm+dm)$ \\
SORF & $O(Nm\log d)$ & $O(Nm)$ & $O(Nm)$ \\
Fastfood & $O(Nm\log d)$ & $O(Nm)$ & $O(Nm)$ \\
RPCholesky & $O(Nmd+Nm^2)$ & $O(Nm)$ & $O(Nm)$ \\
\hline
\end{tabular*}
\vspace{0.5mm}
\begin{minipage}{\columnwidth}
\footnotesize
\emph{Note:} Exact-backend memory is $O(Nd)$ with on-the-fly columns and $O(N^2)$ if the full matrix is stored. RPCholesky's build cost assumes $O(d)$-cost kernel evaluations plus the $O(Nm^2)$ Cholesky update.
\end{minipage}
\end{table}

\begin{table}[!t]
\centering
\footnotesize
\caption{Kernel approximation results on the CIFAR-10 dataset (ratio\,=\,50\%, 10 classes, single seed per class). Mean\,$\pm$\,std AUROC, per-epoch solver fit time, and peak memory at the reported budget~$m$. Pareto-efficient operating points are shown; the exact backend is included as the reference point.}
\label{tab:approx_pareto}
\renewcommand{\arraystretch}{1.1}
\begin{adjustbox}{max width=1\columnwidth}
\begin{tabular*}{\columnwidth}{@{\extracolsep{\fill}}llccc@{}}
\hline
\textbf{Backend} & $m$ & \textbf{AUROC} & \textbf{Fit time (s)} & \textbf{Mem (MB)} \\
\hline
Exact        & --   & $0.982\pm0.016$ & 3.98  & 274.7 \\
\hline
Nystr\"om    & 64   & $0.978\pm0.016$ & 1.44  & 4.0   \\
Nystr\"om    & 256  & $0.977\pm0.016$ & 1.45  & 16.3  \\
RPCholesky   & 128  & $0.973\pm0.021$ & 1.53  & 5.9   \\
QMC-RFF      & 1024 & $0.967\pm0.022$ & 1.56  & 62.9  \\
SORF         & 1024 & $0.966\pm0.032$ & 1.48  & 47.0  \\
Fastfood     & 1024 & $0.962\pm0.028$ & 1.47  & 47.0  \\
Fastfood     & 4096 & $0.976\pm0.017$ & 2.29  & 187.7 \\
ORF          & 2048 & $0.975\pm0.015$ & 2.61  & 125.8 \\
RFF          & 4096 & $0.971\pm0.019$ & 2.41  & 251.6 \\
\hline
\end{tabular*}\end{adjustbox}
\vspace{0.5mm}
\begin{minipage}{\columnwidth}
\footnotesize
\emph{Note:} ORF is included for completeness but is Pareto-dominated at every budget by SORF or Fastfood due to its ${\approx}100{\times}$ higher build cost without a commensurate accuracy gain.
\end{minipage}
\end{table}

\begin{figure*}[!t]
    \centering
    \includegraphics[width=1\textwidth]{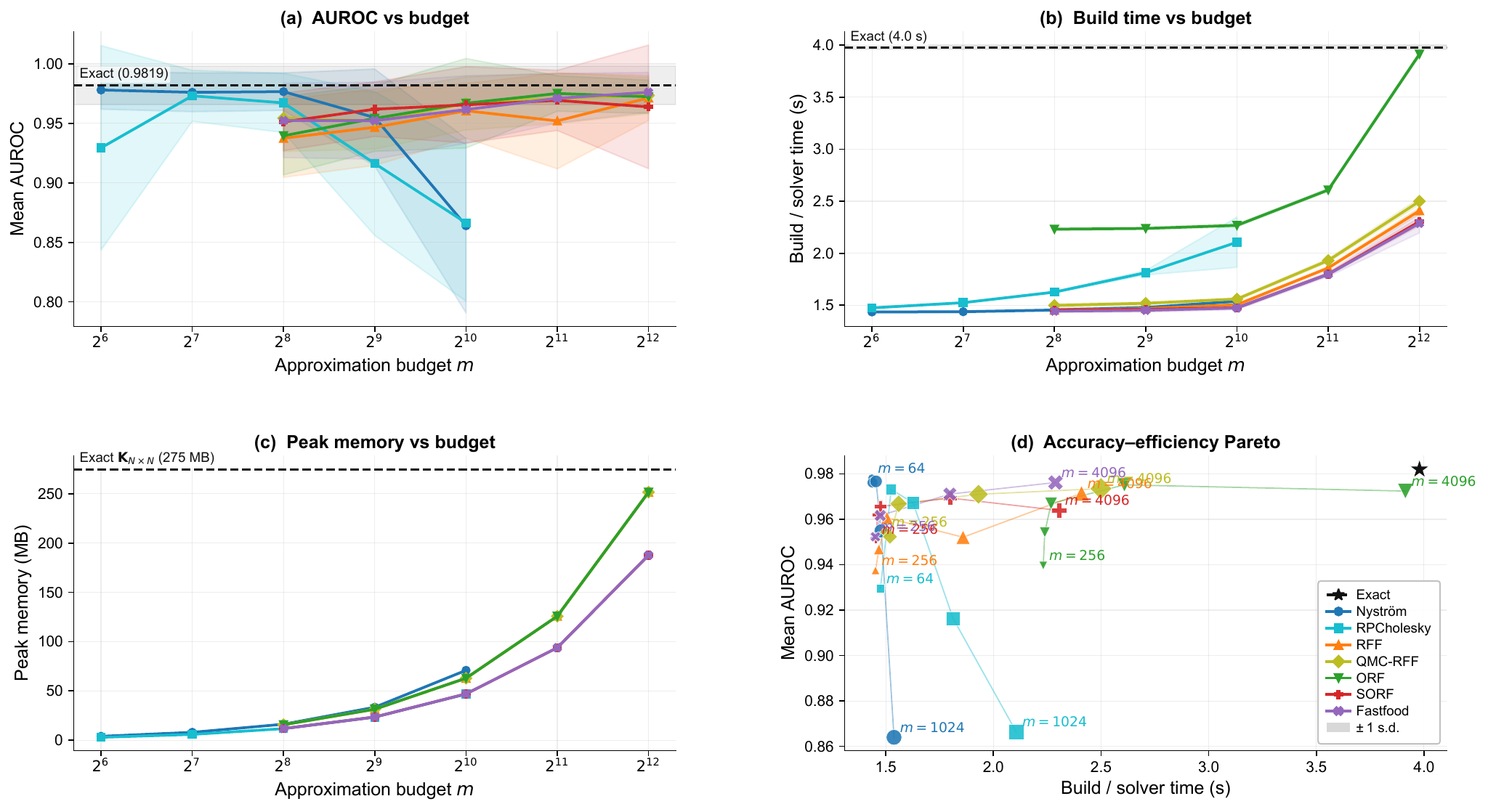}
    \caption{Kernel approximation accuracy--efficiency analysis on CIFAR-10 (negative-to-positive ratio\,=\,50\%; see also Table~\ref{tab:approx_pareto} for selected operating points).
    \textbf{(a)}~Mean AUROC as a function of approximation budget $m$; the dashed line shows the exact-kernel baseline.
    \textbf{(b)}~Build and solver time vs.\ $m$; structured methods (SORF, Fastfood) scale as $O(Nm\log d)$ while dense-projection methods (RFF, ORF, QMC-RFF) scale as $O(Ndm)$.
    \textbf{(c)}~Peak memory vs.\ $m$; the dashed line shows the memory cost of storing the full $\mathbf{K}_{N\times N}$ matrix for the exact backend, whereas approximate backends require only $O(Nm)$ for the explicit feature map $\widetilde{\mathbf{Z}}_{N\times m}$.
    \textbf{(d)}~Accuracy--efficiency Pareto frontier; marker size grows with budget $m$, and the exact kernel appears as a star ($\bigstar$). Shaded bands denote $\pm1$ standard deviation across CIFAR-10 classes.}
    \label{fig:approx_grid}
\end{figure*}

Table~\ref{tab:approx_complexity} summarizes the theoretical complexity of each kernel approximation strategy discussed earlier, while Fig.~\ref{fig:approx_grid} and Table~\ref{tab:approx_pareto} report empirical AUROC, fit time, and memory on the CIFAR-10 dataset (ratio\,=\,50\%) as a function of the approximation budget \(m\). All timing and memory measurements were obtained on an NVIDIA RTX 4090 GPU. The exact-kernel baseline achieves a mean AUROC of \(0.982\) (\(\pm0.016\)), with a peak memory footprint of \(274.7\)\,MB and a per-epoch solver fit time of \(3.98\)\,s, and is used as the reference point for the approximate backends.

To isolate the effect of kernel approximation, all backends are evaluated within the same alternating-training pipeline using identical data splits, backbone initialization, RBF bandwidth, optimization settings, and model-selection protocol. That is, only the approximation method and its budget \(m\) are varied. Accordingly, the reported AUROC, runtime, and memory differences reflect the computational and statistical behavior of the corresponding kernel representation rather than backend-specific retuning. This controlled setting also permits a direct comparison between the empirical trends in Fig.~\ref{fig:approx_grid} and the theoretical complexity estimates in Table~\ref{tab:approx_complexity}.

\textbf{Accuracy vs.\ budget (Fig.~\ref{fig:approx_grid}a).}
Data-dependent approximation strategies reach near-exact accuracy at very small budgets but degrade sharply once $m$ exceeds the kernel's effective rank: Nystr\"om matches the exact kernel to within $0.004$ AUROC at $m\!\le\!256$, then falls to $0.864$ ($\pm0.073$) by $m=1024$; RPCholesky shows the same pattern, peaking at AUROC\,=\,$0.973$ ($m=128$) before degrading similarly. Among spectral methods, accuracy improves monotonically with $m$ but stays below Nystr\"om at equal budgets; QMC-RFF consistently beats standard RFF (up to $+0.017$ AUROC), and Fastfood attains the best spectral AUROC ($0.976$) at $m=4096$. ORF offers no accuracy advantage over SORF/Fastfood despite incurring ${\approx}100\times$ higher build cost, making it Pareto-dominated throughout.

\textbf{Time and memory (Fig.~\ref{fig:approx_grid}b--c).}
Build times follow the theoretical complexity classes: SORF and Fastfood remain sub-millisecond at all budgets ($O(Nm\log d)$), while dense-projection methods (RFF, QMC-RFF, ORF) and RPCholesky grow with $m$. The approximate backends reduce peak memory relative to the exact \(274.7\)\,MB kernel ({\em e.g.}, $69\times$ at Nystr\"om $m=64$), and per-epoch solver time drops by $1.5$--$2.8\times$ across all backends and budgets tested.

\textbf{Pareto frontier (Fig.~\ref{fig:approx_grid}d) and guidance.}
Nystr\"om at $m\!\le\!256$ and RPCholesky at $m=128$ dominate the low-budget regime, combining near-exact accuracy with the lowest time and memory cost. Both, however, are unstable at larger $m$ due to random landmark/pivot selection interacting with the evolving deep features during alternating training. Among spectral methods, Fastfood at \(m=4096\) lies on the Pareto
frontier. In practice, we recommend Nystr\"om or RPCholesky at $m\!\le\!256$ for CIFAR-10-scale problems, switching to Fastfood or QMC-RFF at larger budgets where class-wise stability matters more than peak single-point accuracy.

\subsection{Ablation Study}
\label{subsec:ablation}

\subsubsection{Effect of joint training}
To objectively analyze the merits of joint representation learning, we compare the full alternating framework against a fixed-feature variant in which the CNN backbone is frozen and only the SVDD boundary is learned. Table~\ref{tab:ablation_fixed_cnn} reports results across all three datasets and anomaly ratios for this set of experiments. As may be observed from the table, fixing the CNN leads to a substantial and consistent drop in AUROC compared to the jointly trained model, which confirms that representation adaptation is a key contributor to the observed gains. While the contribution of the proposed joint learning scheme in all cases is positive, as may be observed from the table, the gains from the proposed joint learning scheme may reach as high as \(11.7\%\) in AUROC for some datasets which is a substantial improvement over the baseline.

\begin{table}[!t]
\centering
\footnotesize
\caption{The effect of joint CNN training under the standard one-vs-rest anomaly-detection setting. ``Fixed CNN'' freezes the ImageNet-pretrained backbone and trains only the SVDD boundary, while ``Joint'' refers to the proposed full alternating optimization. Values show average AUROC across 10 random seeds.}

\label{tab:ablation_fixed_cnn}
\begin{adjustbox}{max width=1\columnwidth}
\begin{tabular}{l l c c c}
\hline
\textbf{Dataset} & \textbf{Ratio} & \textbf{Fixed CNN} & \textbf{DLM-SVDD (joint)} & \textbf{$\Delta$ AUROC} \\
\hline
\multirow{3}{*}{F-MNIST}
 & 10\% & 0.902 & 0.976 & +0.074 \\
 & 50\% & 0.901 & 0.987 & +0.086 \\
 & 75\% & 0.901 & 0.988 & +0.087 \\
\hline
\multirow{3}{*}{CIFAR-10}
 & 10\% & 0.867 & 0.977 & +0.110 \\
 & 50\% & 0.868 & 0.982 & +0.114 \\
 & 75\% & 0.867 & 0.984 & +0.117 \\
\hline
\multirow{3}{*}{CIFAR-100}
 & 10\% & 0.923 & 0.962 & +0.039 \\
 & 50\% & 0.923 & 0.971 & +0.048 \\
 & 75\% & 0.923 & 0.972 & +0.049 \\
\hline
\end{tabular}\end{adjustbox}
\end{table}

\subsubsection{Effect of kernel approximation}
Section~\ref{subsec:approx_results} detailed the accuracy--efficiency behavior of the seven kernel backends considered within the Frank--Wolfe boundary-fitting step. From an ablation standpoint, the key observation is that the choice of approximation strategy is not a minor implementation detail: swapping the exact kernel for a poorly matched backend or budget can cost several points of AUROC, whereas a well-matched data-dependent backend (Nystr\"om or RPCholesky) at a small budget recovers essentially the same accuracy as the exact kernel at a fraction of the memory and time. This confirms that the large-margin boundary itself is robust to the specific kernel
representation, provided the approximation rank is adequate for the target dataset, and it validates treating the backend and budget as tunable deployment parameters rather than fixed choices baked into the method.

\subsubsection{Effect of RBF kernel bandwidth}
In this section, we study the effect of different strategies to estimate the RBF kernel bandwidth $\gamma$, comparing an arbitrary fixed value chosen before training, re-estimation from the evolving deep features at the start of every epoch, and a one-time estimate from the pretrained ImageNet features based on the median squared pairwise Euclidean distance (used throughout this work). Based on our analysis, the arbitrary fixed choice was observed to be the weakest of the three, as expected given it ignores the actual feature scale. More notably, continual re-estimation from the evolving features was not the best strategy either: as deep features are themselves being updated toward the current kernel scale in the $\omega$-step, repeatedly re-fitting $\gamma$ to these same features compounds with that adaptation and encourages the representation to overfit to a moving kernel target, which was observed to lead to increasingly unstable validation performance across epochs. Estimating $\gamma$ once from the pretrained features and holding it fixed avoids this feedback loop: the bandwidth reflects a stable, representation-independent notion of scale, so the alternating optimization is left with a single degree of freedom (the CNN) to adapt to the boundary, rather than two interacting and mutually reinforcing sources of drift.

\subsubsection{Effect of within-task anomaly supervision}
We next examine how much labeled-anomaly supervision is needed \emph{within a single one-vs-rest task}. Here the negative-to-positive ratio shows the number of labeled anomalous training samples relative to the number of normal training samples (10\%, 50\%, or 75\%), while the normal class remains fully represented. As reported in Table~\ref{tab:overall_results}, DLM-SVDD remains competitive even at the lowest supervision level and improves further as more anomalous samples are exposed. This is in line with the large-margin formulation, as more anomalous samples become available, optimization can benefit from such additional sources of information, rather than relying solely on the compactness of the normal region. Therefore, additional anomaly supervision yields diminishing yet positive returns.

\subsubsection{Sensitivity to training-population imbalance}
A separate question concerns how the framework behaves when the underlying multi-class pool used to construct the one-vs-rest tasks is itself long-tailed, independent of the within-task ratio above. Following the construction in Eq.~\eqref{eq:lt_formula}, class cardinalities decay exponentially from head to tail as controlled by $\varrho \in \{10, 50, 100\}$, with $\varrho = 100$ denoting the most severe skew. Table~\ref{tab:lt_results} reports balanced accuracy on the CIFAR-10-LT and CIFAR-100-LT datasets under this protocol. As $\varrho$ increases, jointly trained DLM-SVDD shows the expected decline in balanced accuracy, yet its margin over the fixed-CNN variant \emph{widens} rather than shrinks as the imbalance worsens---on the CIFAR-10-LT dataset, for instance, the gain over the fixed-CNN counterpart grows to 33.75, 23.93, and 27.29 balanced-accuracy points at $\varrho = 10, 50, 100$, respectively. Adapting the representation to the target distribution therefore becomes more important as class frequencies become more skewed, since a fixed pretrained feature space is progressively less able to compensate for severely imbalanced training populations.

\subsection*{Reproducibility}
To facilitate reproducibility, the source code, training scripts, and  instructions for reproducing all reported results are made publicly available at \url{https://github.com/AlirezaSaei1/DLM-SVDD}.

\section{Conclusion}
\label{sec:conclusion}

We presented DLM-SVDD, a deep large-margin $\ell_p$-SVDD framework that jointly learns convolutional neural network representations and an explicit margin-aware SVDD boundary through a novel alternating optimization scheme. The $\alpha$-step solves the large-margin $\ell_p$-SVDD dual problem via the Frank--Wolfe algorithm subject to fixed deep features, while the $\omega$-step updates the CNN backbone parameters through a smooth margin-violation surrogate loss that focuses gradient signal on decision-critical, margin-violating samples. Experiments on standard one-vs-rest anomaly-detection benchmarks demonstrate consistent improvements over the fixed-feature predecessor and yield a superior performance compared with state-of-the-art methods. Under severe long-tailed training distributions conditions, the proposed joint model substantially improves over the fixed-CNN baseline and achieves the best average rank compared to other methods from the literature.

To address scalability, we introduced and systematically evaluated a family of kernel approximation strategies within the proposed alternating optimization approach. Data-dependent methods (Nystr\"{o}m and RPCholesky) achieved near-exact detection accuracy at small approximation budgets with significant memory and time reductions, while structured spectral methods (SORF and Fastfood) offer favorable trade-offs at larger budgets.

\bibliographystyle{IEEEtran}
\bibliography{refs}

\end{document}